\documentclass[journal]{IEEEtran}

\usepackage{amssymb}
\usepackage{amsmath}
\usepackage{xcolor}
\usepackage{hhline}
\usepackage{multirow}
\usepackage{booktabs}
\usepackage{graphicx}

\begin{document}

\title{Mapping Woody Vegetation from Multi-Source Imagery and Prediction Fusion for Enhanced Data Efficiency and Accuracy}

\author{Kal~Backman,
        Jared~Wood and
        Adam~Roff
\thanks{Corresponding author: Kal Backman}
\thanks{The authors are with the New South Wales Department of Climate Change, Energy, the Environment and Water, Parramatta, 2150, NSW, Australia. Email: [Kal.Backman, Jared.Wood, Adam.Roff]@dcceew.nsw.gov.au}
\thanks{K. Backman is also affiliated with Monash University, Clayton, 3800, Vic, Australia. Email: Kal.Backman@monash.edu}
\thanks{A. Roff is also affiliated with Earth Observation Lab, University of New England, Armidale, 2351, NSW, Australia and the Conservation Science Research Group, University of Newcastle, Callaghan, 2308, NSW, Australia}}

\markboth{Journal of \LaTeX\ Class Files,~Vol.~X, No.~X, Month~20XX}%
{Shell \MakeLowercase{\textit{et al.}}: Bare Demo of IEEEtran.cls for Journals}
%



\maketitle

\begin{abstract}
Tree cover maps are a fundamental remote sensing product, used to derive ecological insights about the landscape and are essential to change detection, vegetation mapping and fire monitoring programs. 
However, comprehensive tree cover mapping requires reliable and high-quality imagery, free of cloud and weather defects to ensure accurate model outputs.
Deep learning approaches can generate high quality maps with minimal human intervention but require large amounts of human annotated data to be successful.
In this work we propose a framework consisting of methods that aim to improve the data efficiency and robustness of deep learning models using data fusion techniques to segment woody vegetation defined as vegetation over the height of 2m across the state of New South Wales, Australia.
To improve robustness against varying image quality, we propose an image composition method that normalizes the imagery and removes defects, whilst also minimizing the reliance on individual image quality by proposing a prediction fusion method. The two methods resulted in an error reduction of 38.2\% and 53.6\% respectively compared to single-source imagery. 
To address deep learning approaches’ limitation of requiring large amounts of data, we apply label transfer to multiple sources of imagery as a form of data augmentation to improve data efficiency. Learning from multiple image sources was shown to be the biggest improvement in performance, resulting in an error reduction between 28.1\% to 76.2\% across the different validation experiments, whilst reducing the standard deviation of performance across image dates by a factor of 13.
\end{abstract}


\IEEEpeerreviewmaketitle

\section{Introduction}\label{Section_Introduction}
Tree cover maps play a quintessential role in remote sensing operations, used in environmental monitoring to derive carbon sequestration sources \cite{patra2013carbon, yang2022terrestrial, artaxo2022tropical} and for habit monitoring and fragmentation analysis \cite{haddad2015habitat, fisher2016large, newbold2015global}.
Such tree cover maps act as a foundational remote sensing product, used as a predictor, error correction and spatial constraint mask in change detection \cite{scarth2008assimilation, qdec2018slats, potapov2011regional}, vegetation mapping \cite{lourencco2021estimating, pu2012comparative} and fire monitoring \cite{gibson2020remote, hislop2023using}.

Traditional tree cover mapping approaches focus on establishing a set of predictors deriving from various spectral bands and indices to be used in modelling techniques such as thresholding \cite{pu2012comparative, boggs2010assessment, qin2016mapping}, logistic regression \cite{fisher2016large}, regression trees \cite{erker2019statewide, yang2019fractional, potapov2017comprehensive} and support vector machines \cite{eskandari2020mapping, erker2019statewide, mironczuk2017mapping}. Such approaches excel in terms of data efficiency, requiring a relatively small sample size of easily to obtain isolated points. However such traditional approaches are limited in their ability to infer spatial context cues due to being constrained to predefined texture metrics with limited contextual fields, resulting in spatial coherency issues denoted as speckled holes and isolated predictions \cite{huang2004post, liu2024coastal, bontemps2008object}.

Conversely, deep learning approaches demonstrate strong spatial consistency in predictions due to employing convolutional layers at varying spatial scales to pool information spatially. Such deep learning approaches tend to use variations of the U-Net architecture \cite{morford2024mapping, flood2019using, chen2023urban} to map vegetation. However deep learning approaches’ greater performance \cite{sothe2020comparative, boston2022comparing, liu2024coastal, onojeghuo2023wetlands, liu2018comparing} comes at the cost of requiring orders of magnitude more examples compared to non-deep learning approaches as seen in Table~\ref{Table_priorwork}. 

\begin{table*}
\caption{Labels required by study area for woody vegetation mapping}
\hskip 1.25cm
\scriptsize
\label{Table_priorwork}
\begin{tabular}{l|lrrr}
\bottomrule
Reference & Method & Labeled training pixels & Training study area (km\(^2\)) & Training pixels / area \\
\toprule
\hline
\cite{chen2023urban} Chen et al. (2023) & CNN & 46,923,776 & 65 & 721,904.25 \\
\cite{lin2023urban} Lin et al. (2023) & CNN & 160,432,128 & 285 & 561,599.50 \\
\cite{freudenberg2022individual} Freudenberg et al. (2022) & CNN & 29,333,304 & 250 & 173,333.22 \\
\cite{morford2024mapping} Morford et al. (2024) & CNN & 8,960,000,000 & 193,000 & 46,424.87 \\
\cite{he2022generating} He et al. (2022) & CNN & 566,476,800 & *179,700 & 3,152.35 \\
\cite{morford2024mapping} Morford et al. (2024) & CNN & 432,000,000 & 193,000 & 2,238.34 \\
\cite{cheng2023dual} Cheng et al. (2023) & CNN & 50,462,720 & *53,000 & 952.13 \\
\cite{flood2019using} Flood et al. (2019) & CNN & 578,900,000 & 1,730,000 & 334.62 \\
\cite{barnetson2019mapping} Barnetson et al. (2019) & Random forests & 13,102 & *41 & 319.56 \\
\textbf{Ours} & \textbf{CNN} & \textbf{93,896,413} & \textbf{801,137} & \textbf{117.20} \\
\cite{liao2020woody} Liao et al. (2020) & Random forests & 40,585 & 3,431 & 11.83 \\
\cite{potapov2017comprehensive} Potapov et al. (2017) & Decision tree & 640,000 & 149,387 & 4.28 \\
\cite{noelke2021continuous} Noelke (2021) & Multi-layer perceptron & 880 & 250 & 3.52 \\
\cite{gonzalez2016tree} González-Roglich and Swenson (2016) & Random forests & 150,000 & 50,000 & 3.00 \\
\cite{mironczuk2017mapping} Mirończuk and Hościło (2017) & Support vector machine & 240 & 168 & 1.43 \\
\cite{fisher2016large} Fisher et al. (2016) & Logistic regression & 26,579 & 809,444 & 0.03 \\
\cite{higginbottom2018mapping} Higginbottom et al. (2018) & Random forests & 1,800 & 125,000 & 0.01 \\

\hline
\bottomrule
\end{tabular}
\newline
Table entries prefixed with * indicate values which required additional external information or additional analysis from figures within the respective works. For cases in which labeled training pixels were not explicitly reported, they were derived from the reported training tiles, tile resolution and pixel resolution. Prior works are ranked from least to most efficient based on the overall workflow efficiency of labels required normalized by training study area coverage. Training study area coverage is defined as the approximate convex area of training samples, not the explicit pixel area coverage of training data to account for differences in image resolution and breadth of vegetation modeled.
\end{table*}

Further issues experienced by both methods is the reliance on high quality remote sensing imagery that is free of cloud, shadow artifacts and weather events.
Developing comprehensive vegetation maps across large study areas from a single imagery date is difficult due to the prevalence of clouds obstructing the view of the earth’s surface. Methods to mitigate such image defects utilize cloud masks \cite{potapov2011regional} to flag poor quality images needing to be recaptured and composited or the use of time series imagery \cite{fisher2016large} to minimize the impact of cloudy data sources. However cloud artifacts are not the only source of image degradation, lighting conditions and shadow length deriving from sun elevation and weather events like flooding and snow further degrade image quality. Making attaining a consistent and high quality visual representation of the landscape difficult, often requiring manual intervention to ensure quality.

In this work we aim to develop a deep-learning framework to detect woody vegetation, which we define as vegetation over the height of 2 meters. The framework comprises of a convolutional neural network (CNN) encoder-decoder network in a U-Net \cite{ronneberger2015u} configuration that outputs a segmentation mask denoting woody vegetation. The framework’s core objectives are to enhance data efficiency in order to address deep learning approaches’ main limitations, and to improve robustness to ensure consistency in model outputs against vary image quality.

To enhance data efficiency, we transfer partially labelled images to multiple image sources as a form of data augmentation. To enhance robustness, we propose a multi-date image composition method which requires no additional input masks or manual intervention steps in order to reduce the susceptibility of image defects resultant from clouds and shadows. For greater robustness against varying image quality, a multi-image prediction fusion method is proposed to improve the framework's ability to generate accurate woody vegetation maps.

The main contributions of the proposed work are:
\begin{enumerate}
\item The development of a deep-learning framework for woody vegetation segmentation that maps the entire state of New South Wales, Australia covering 801,137 km\(^2\) of land area.
\item Label transfer to multiple image sources as a form of data augmentation for increased data efficiency and performance. When compared to prior deep learning works, it was shown to result in the most efficient model on a label per study area basis, whilst capable of reducing the error by up to 76.2\% in validation experiments.
\item A multi-image composition method and a multi-image prediction fusion method for increased accuracy and robustness against image defects, which when coupled resulted in an error reduction of 53.6\%.
\end{enumerate}

\section{Methodology}
\subsection{Data}\label{Section_Methodology_Data}
\subsubsection{Raw input imagery}
The input imagery used for the woody segmentation framework is derived from 1.5m pansharpened SPOT 6/7 imagery which comprised of 4 bands: blue (454-519nm), green (527-587nm), red (624-694nm) and NIR (756-880nm). 
SPOT imagery was captured yearly for a total of 4 years across the 2019 to 2022 period where each yearly capture was performed across mainland New South Wales, Australia, covering a total landmass of 801,137 km\(^2\) per year.
The SPOT imagery was split into 343 roughly 50km \(\times\) 50km scenes which could be stacked between subsequent years. 

\subsubsection{Multi-date image composition}
As relying on single date imagery is susceptible to cloud, shadow and lighting artifacts, a composite of multi-date SPOT imagery was created to normalize image quality. The multi-date composite was created by taking the discrete-median of the 4 yearly images across each of the 4 bands, where each yearly image was normalized prior to the median operation to account for difference in lighting conditions. The approach requires no additional masks to remove cloud artifacts nor does it require manual intervention to set normalization parameters.

Image normalization was performed by fitting a single and double gaussian distribution to each image band histogram. The gaussian model chosen to normalize the image band was determined by the model containing the lower sum of squared errors between the model and the computed band histogram.
Each image band was subsequently divided by the mean of the gaussian distribution to normalize the image band. For the double gaussian distribution which contained two mean values, the lower of the two means was used given that the frequency count of the higher mean did not exceed that of the lower mean by a threshold factor of 5.
The lower mean value was chosen to better emphasize woody vegetation features which appear darker with respect to the imagery whilst the threshold factor check was used to account for small spikes from scenes containing many bodies of water.
The combination of single and double gaussian methods were used due to the large variability of cloudiness within the imagery, where a single gaussian catered towards little cloud cover whilst the double gaussian aimed to normalize the non-cloudy values for cloud abundant imagery.

Given each of the 4 normalized images, a median operation was performed where each chosen value from the median operation resulted in a discrete pixel value being chosen from one of the normalized bands. The median operation did not perform an averaging operation to determine the median value for an even set of values but instead picked the lower of the two values such that each pixel could be associated with a specific image source. 

\begin{figure}
\centering
\includegraphics[width=1.0\columnwidth]{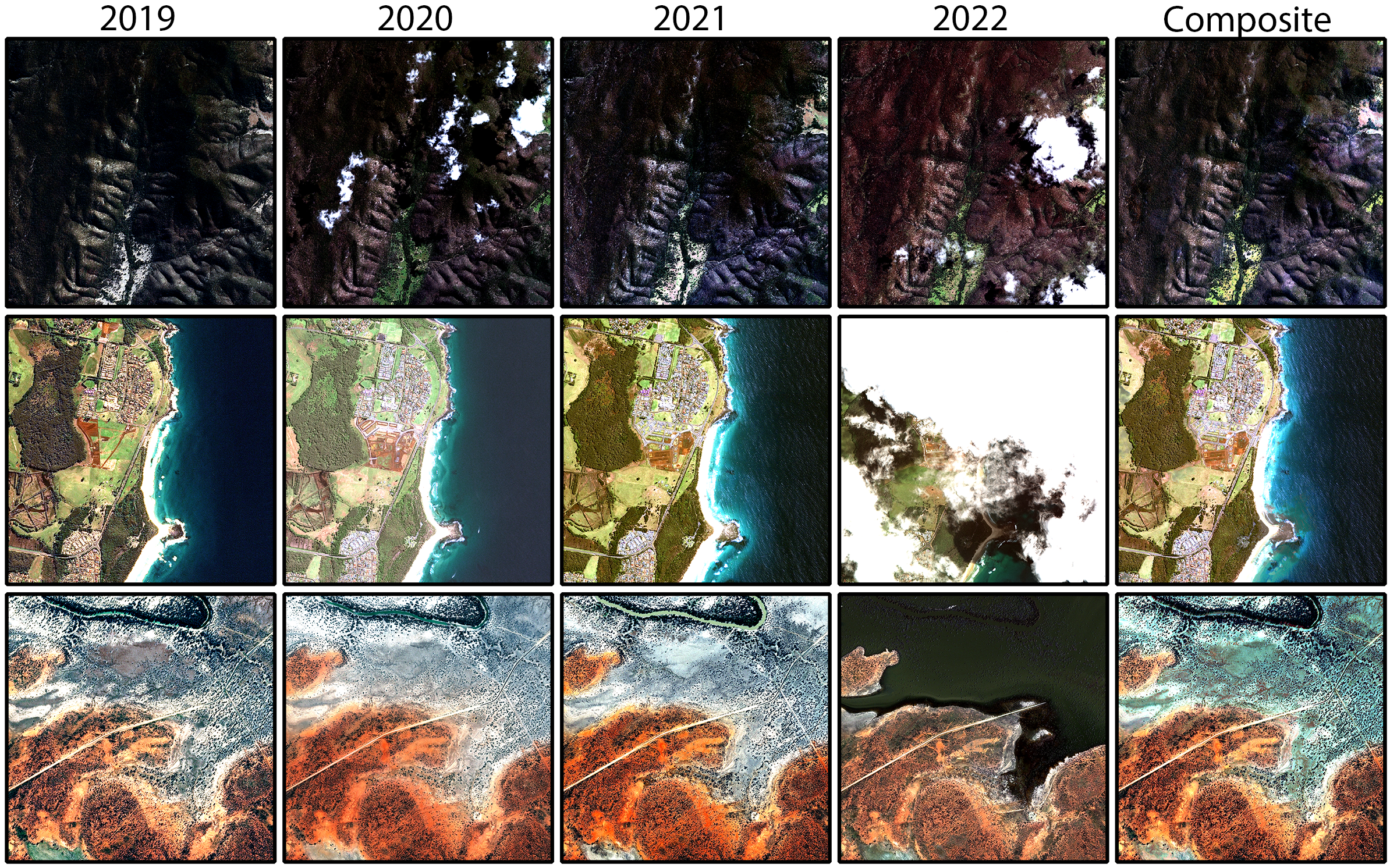}
\caption{Examples of the individual SPOT images used to create the SPOT image composite.
Despite the presence of cloud and flooding events, the image composite offers a more standardized source of imagery.}
\label{SPOTcomposition}
\end{figure}

Examples of the SPOT composite imagery and the individual date SPOT images used to create the composition can be found in Fig.~\ref{SPOTcomposition}.

\subsubsection{Woody extent labels}
To generate labels required to train the model, 5 of the 343 scenes were initially selected from across the study area. For each of the 5 selected scenes, a total of 80 non-overlapping randomly sampled image patches of size 512 \(\times\) 512 pixels were extracted from the multi-date image composite. For each extracted scene, operators were tasked with assigning a binary label denoting if a pixel contains woody vegetation over the height of 2m or not. To supplement the labeling procedure, auxiliary data sources derived from Google Earth and Google Street View were used by operators to help disambiguate the SPOT composite imagery.
Operators were required to cross reference each geotagged 512 \(\times\) 512 image patch against higher resolution Google Earth imagery, utilizing multiple image dates to help infer vegetation growth and species from seasonality changes.
Where available, Google Street View imagery was favored as an auxiliary source for assessing vegetation height in image patches. For image patches where street view imagery was non-existent, operators were encouraged to use street view imagery of neighboring regions sharing similar vegetation photo patterns to cross reference photo patterns with vegetation species.

Due to the ambiguity of accurately discerning if woody vegetation is over the height of 2m, and the difficulty of ensuring completeness in the labels for image patches containing large quantities of isolated tree crowns, an additional weighting mask was introduced. Operators were able to assign pixel values to the weight mask to reflect their confidence in the labels in order to minimize the impact of potential erroneous labels \cite{backman2024bicycle} due to the ambiguity of used data sources, and to facilitate the partial completion of an image patch. Assigning pixel values of 0 to the weight mask resulted in the corresponding label pixels being ignored whilst the remaining values were used to scale the loss function as described in Section~\ref{section_Training}.

To attain a greater diversity of training data and to more efficiently improve model performance, additional training samples were generated by targeting data samples that exhibit poor model performance. 
An additional 10 scenes that were uniformly distributed across the study area were chosen. A model trained as outlined in Section~\ref{section_Training} was used to generate a confidence mask by fusing multiple SPOT image sources as described in Section~\ref{section_inference}. 

The generated model confidence mask was used to propose regions of uncertainty by tiling the scene into 512 \(\times\) 512 confidence patches. A confidence threshold was applied and a patch uncertainty score (\(U_{score}\)) was computed by performing a weighted sum of threshold confidence values:
\begin{equation}
 U_{score} = \sum\limits_{i=1}^{N}{2 {p_i} (\hat{y_i} < T)} + \sum\limits_{i=1}^{N}{(1-{p_i}) (\hat{y_i} < T)},
\end{equation}
where \({p_i}\) denotes the predicted class at the \(i\)th pixel which takes a value of 1 for a predicted woody pixel and a value of 0 for a non-woody pixel and is derived from thresholding the corresponding confidence value \(\hat{y_i}\). \(T\) is a confidence threshold constant. Woody predicted pixels are weighted by a factor of 2 to prioritize regions that contain more woody vegetation.

Image patches were sorted by their uncertainty score where the 100 most uncertain image patches per scene were denoted as patches requiring labeling given that the uncertainty score was above a specified threshold. Operators were assigned proposed image patches accompanied by the model’s predictions as pre-labels. Operators were encouraged to select image patches where the model’s predictions were in the most in need of correction and were given the option to skip patches. 

The process of training a new model and proposing new image patches of uncertainty to label from unseen scenes was repeated until satisfactory model performance. A total of 1,365 labeled image patches were collected which derived from a total of 86 scenes spread across the study area.

\subsubsection{Training and validation dataset}
A dataset used to train and validate the model was generated by initially normalizing the input imagery bands to the range of [-1, 1]. A histogram for each band across each scene was computed, where the lowest and highest 2.5\% of values were clipped and rescaled to the range of [-1, 1]. 
To maximize data efficiency, the dataset contained multiple input image sources which included the individual yearly SPOT 6/7 imagery from 2019 to 2022 and the SPOT image composite. The SPOT composite labels were transferred to the individual SPOT image dates by directly overlaying the two. To reduce the incorrect assignment of the woody class for single date SPOT imagery resultant from cloud cover, a cloud mask was used to zero the weight mask for individual images only. Image patches corresponding to a weight mask sum below a threshold value were excluded due to not containing sufficient valid pixels. Transferring the manually annotated labels allowed for a total of 5 data points to be generated for the human labor cost of 1.   

\begin{table*}
\caption{Dataset summary}
\label{TableDataset}
\hskip 3.0cm
\scriptsize
\begin{tabular}{|l|c|c|c|c|c|c|}
\hline
 & \multicolumn{3}{|c|}{\textbf{Training dataset}} & \multicolumn{3}{|c|}{\textbf{Validation dataset}} \\
 \hline
 Image source & Woody pixels & Non-woody pixels & Images & Woody pixels & Non-woody pixels & Images \\
\hline
SPOT composite & 30,079,919 & 63,816,494 & 774 & 21,710,150 & 42,709,484 & 591 \\ 
SPOT 2019 & 28,267,582 & 58,954,508 & 739 & 20,095,319 & 39,808,564 &  557 \\ 
SPOT 2020 & 18,578,928 & 60,558,074 & 697 & 15,487,538 & 41,727,845 &  549 \\ 
SPOT 2021 & 25,954,916 & 51,902,199 & 691 & 19,143,388 & 35,143,642 &  526 \\ 
SPOT 2022 & 27,668,239 & 52,895,068 & 677 & 20,038,349 & 35,275,269 &  502 \\
\hhline{|=|=|=|=|=|=|=|}
Total & 130,549,584 & 288,126,343 & 3,578 & 96,474,744 & 194,664,804 & 2,725 \\ 
\hline
\end{tabular}
\newline
Summary of the training and validation datasets. The total woody and non-woody pixels only include pixels that contain a weighting greater than zero.
\end{table*}

The dataset was probabilistically split such that approximately 55\% of image patches were assigned for training whilst the remaining 45\% of image patches were reserved for validation. For scenes with greater than 2 samples, it was guaranteed that at least 1 sample was assigned for validation. Image patches reserved for validation were applied across all input image sources, ensuring that validation patches from the 2019 imagery would not have their corresponding image patch from 2020 appear in the training dataset. A summary of the training and validation dataset can be found in Table~\ref{TableDataset}.

\subsection{Model architecture}
\begin{figure}[b]
\centering
\includegraphics[width=1.00\columnwidth]{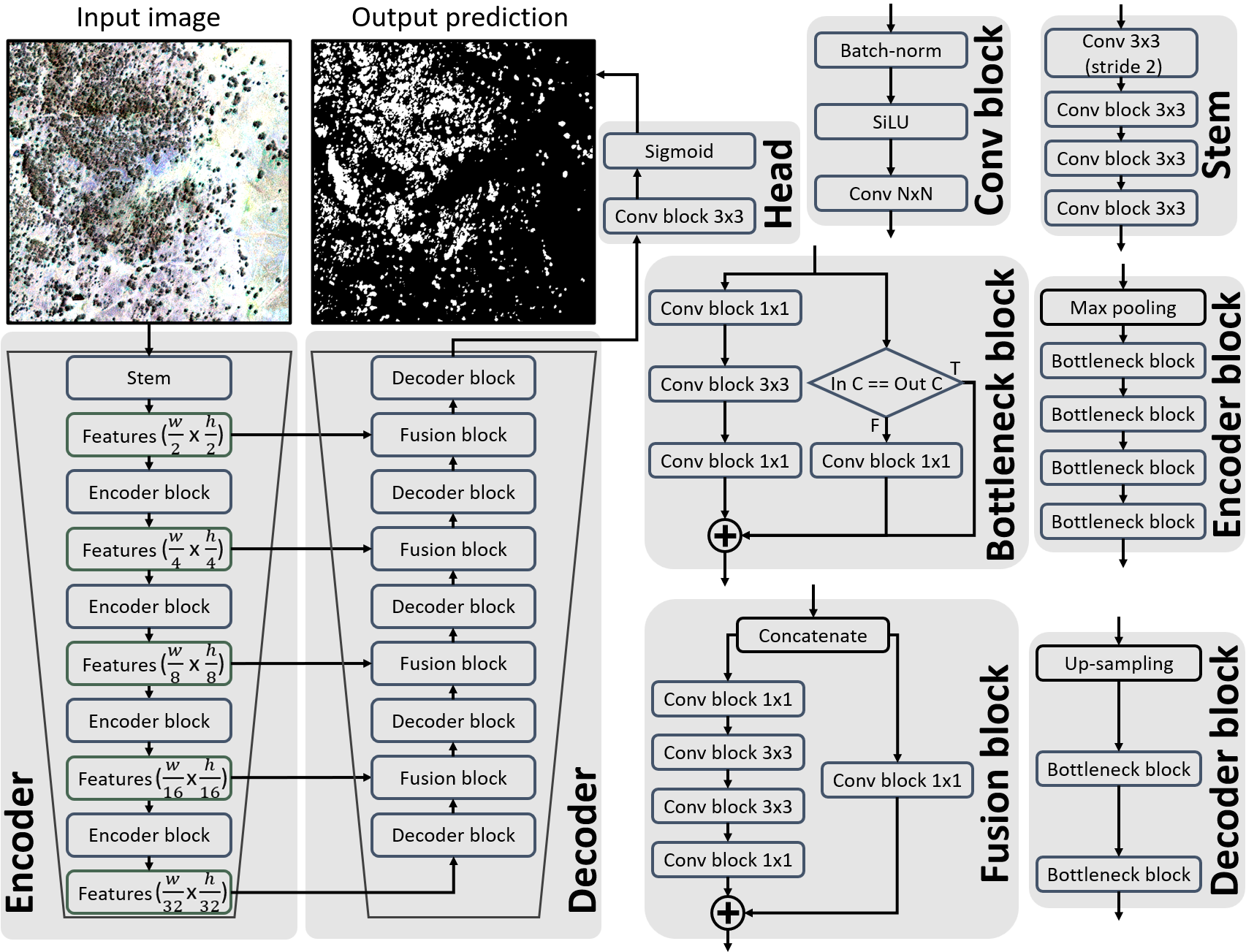}
\caption{Overview of the woody vegetation segmentation model architecture}
\label{modelOverview}
\end{figure}

The model architecture used to segment woody vegetation comprises of an encoder, decoder and head in a U-Net \cite{ronneberger2015u} configuration. An overview of the model architecture can be seen in Fig.~\ref{modelOverview}.

The encoder receives the normalized SPOT imagery and outputs a total of 5 feature maps containing [64, 128, 256, 512, 1028] channels at resolution [\(\frac{1}{2}\), \(\frac{1}{4}\), \(\frac{1}{8}\), \(\frac{1}{16}\), \(\frac{1}{32}\)] relative to the input imagery.
The initial feature map is generated by the encoder’s stem which applies an initial convolutional layer with stride length of 2, followed by 3 consecutive convolution blocks comprising of batch-norm, SiLU activation and convolutional layers. 
Subsequent feature maps are generated by the encoder block which comprises of an initial max pooling operation followed by 4 consecutive bottleneck residual blocks \cite{he2016deep}.

The decoder receives the 5 image feature maps from the encoder and applies a decoder block comprising of an initial up-sampling operation followed by 2 consecutive bottleneck blocks to the deepest feature map.
A fusion block is then applied which concatenates the output from the prior decoder block with that of the feature map from the encoder that corresponds to the current feature resolution. 
The process of applying a decoder block to up-scale the features and the fusion of features from the encoder is repeated until the resolution of the feature map matches the input image resolution.

The head applies a single convolution block to the output of the decoder, followed by a sigmoid activation function to generate confidence scores that a pixel within the output image contains woody vegetation.

\subsection{Training}\label{section_Training}
The woody vegetation segmentation model was trained by randomly sampling from the data-weight-label pairs derived from the SPOT image composite and individual date SPOT image patches using a batch size of 80. 
The sampled training image patches had a stochastic data augmentation process applied to them to further diversify training examples. Data augmentation techniques include scaling transforms such as gamma corrections to artificially darken or brighten the image as well as band-shift transforms to make individual bands more prominent.
Noising transforms were implemented by adding normally distributed noise and gaussian blur operations to distort the input imagery. 
Geometric transforms were used which include axis flipping, image rotation, scaling and shear operations to alter the perspective and shape of the input imagery.
The augmented data-weight-label pairs subsequently had a 416 \(\times\) 416 chunk extracted from the 512 \(\times\) 512 image patch to be used as input to the model. The 416 \(\times\) 416 chunk was randomly chosen whilst constrained to minimize the area of invalid data pixels resultant from geometric transforms deforming the 512 \(\times\) 512 image patch. 

The model was trained for a total of 160,000 optimization iterations split into 80 epochs containing 2,000 optimization iterations. The loss function used was a weighted sum between the binary cross entropy loss and dice loss:
\begin{equation}
{L_{Total} = w_{BCE} \cdot L_{BCE} + w_{Dice} \cdot L_{Dice}},
\end{equation}
where the binary cross entropy loss is defined as \begin{equation}\label{Equation_weightedBCE}
L_{BCE} = -\frac{1}{N}\sum\limits_{i=1}^{N}( y_i \log(\hat{y_i}) + (1 - y_i) \log(1-\hat{y_i})) w_i  
\end{equation}  
and the dice loss defined as 
\begin{equation}\label{Equation_weightedDice}
L_{Dice} = 1 - \frac{2 \sum\limits_{i=1}^{N} y_{i} \hat{y_i} w_i + \epsilon}{\sum\limits_{i=1}^{N} y_i w_i + \sum\limits_{i=1}^{N} \hat{y_i} w_i + \epsilon}.
\end{equation} 
Weighting coefficient \(w_i\) as seen in Equations~\ref{Equation_weightedBCE} \& \ref{Equation_weightedDice} was derived from the weight mask used to scale the loss function based on the operator’s confidence in label \(y_i\) and to ignore unlabeled pixels. 
\(\epsilon\) is used in Equation~\ref{Equation_weightedDice} for numerical stability to prevent small denominator division errors.  
Binary cross entropy and dice loss scaling terms \(w_{BCE}\) and \(w_{Dice}\) are assigned a value of 0.2 and 1.0 respectively.

The Adam optimizer was used with an initial learning rate of 2e-4 which was linearly decayed to the final learning rate of 1e-5. Gradient clipping was performed using a maximum gradient norm of 1.0 and gradient scaling for mixed precision layers. The model was trained using a NVIDIA A100 80GB GPU.

\subsection{Inference \& prediction fusion}\label{section_inference}
To generate woody extent predictions for a given scene, the model received an input image size of 1024 \(\times\) 1024 using a batch size of 32. The 1024 \(\times\) 1024 image patch was swept across the scene imagery using an image overlap of 50\%. Overlapping output prediction confidence values were fused by taking a weighted average where confidence values closer to the center of the image patch were weighted higher than confidence values residing around the edges of the image patch.

To leverage information across multiple image sources, the predicted confidence masks generated from the single-date and SPOT composite imagery were fused. 
Prediction fusion was performed by initially remapping confidence scores \(\hat{y_i}\) such that a value of 0 denotes a low confidence whilst a value of 1 denotes a high confidence irrespective of the predicted class: 
\begin{equation}
\hat{y}^\prime_i = 
    \begin{cases}
        (\frac{\hat{y_i} - 0.5}{0.5})^2 &\text{if \(\hat{y_i} > 0.5\)} \\
        (\frac{0.5 - \hat{y_i}}{0.5})^2 &\text{else}
    \end{cases}.
\end{equation}

The predicted class was subsequently remapped where a predicted pixel denoting woody vegetation was encoded as 1 whilst non-woody vegetation was encoded with a value of -1:
\begin{equation}
{p}^\prime_i = 
    \begin{cases}
        1 &\text{if \(\hat{y_i} > 0.5\)} \\
        -1 &\text{else}
    \end{cases}.
\end{equation}

A predicted class consistency coefficient was calculated by taking a weighted average across the image sources using the product between the remapped predicted class and confidence:
\begin{equation}
\hat{p_i} = \frac{\sum\limits_{j=1}^{N}{{p}^\prime_{ij} \hat{y}^\prime_{ij} w_{ij}}}{\sum\limits_{j=1}^{N}{w_{ij}}},
\end{equation}
where index \(i\) denotes the \(i\)th pixel from image source of index \(j\).
The corresponding weight for pixel \(i\) of image source \(j\) (\(w_{ij}\)) is derived from the cloud mask used to mask single-date SPOT images with an additional scaling factor of 2 applied to the SPOT composite images due to a greater confidence in the quality of the composite image source.
The class consistency coefficient \(\hat{p_i}\) takes the range of [-1, 1] where a value of -1 denotes a high consistency of high confidence non-woody pixels whilst a value of 1 denotes a high consistency of high confidence woody vegetation pixels.

A binary classification for pixel \(i\) is subsequently assigned by thresholding the class consistency coefficient:
\begin{equation}
\bar{p_i} = 
    \begin{cases}
        1 &\text{if \(\hat{p_i} > 0 \) } \\
        0 &\text{else}
    \end{cases}.
\end{equation}

Confidence value for pixel \(i\) is calculated by taking a weighted average of the remapped confidence values across image sources multiplied by the class consistency coefficient squared.  
\begin{equation}
\bar{y_i} = \frac{\sum\limits_{j=1}^{N}{\hat{y}^\prime_{ij} w_{ij}}}{\sum\limits_{j=1}^{N}{w_{ij}}} \hat{p_i} ^ 2
\end{equation}
The output confidence value \(\bar{y_i}\) uses the class consistency coefficient to greater decay confidence values where there is a disagreement between the predicted class between multiple image sources.

\begin{figure}
\centering
\includegraphics[width=1.0\columnwidth]{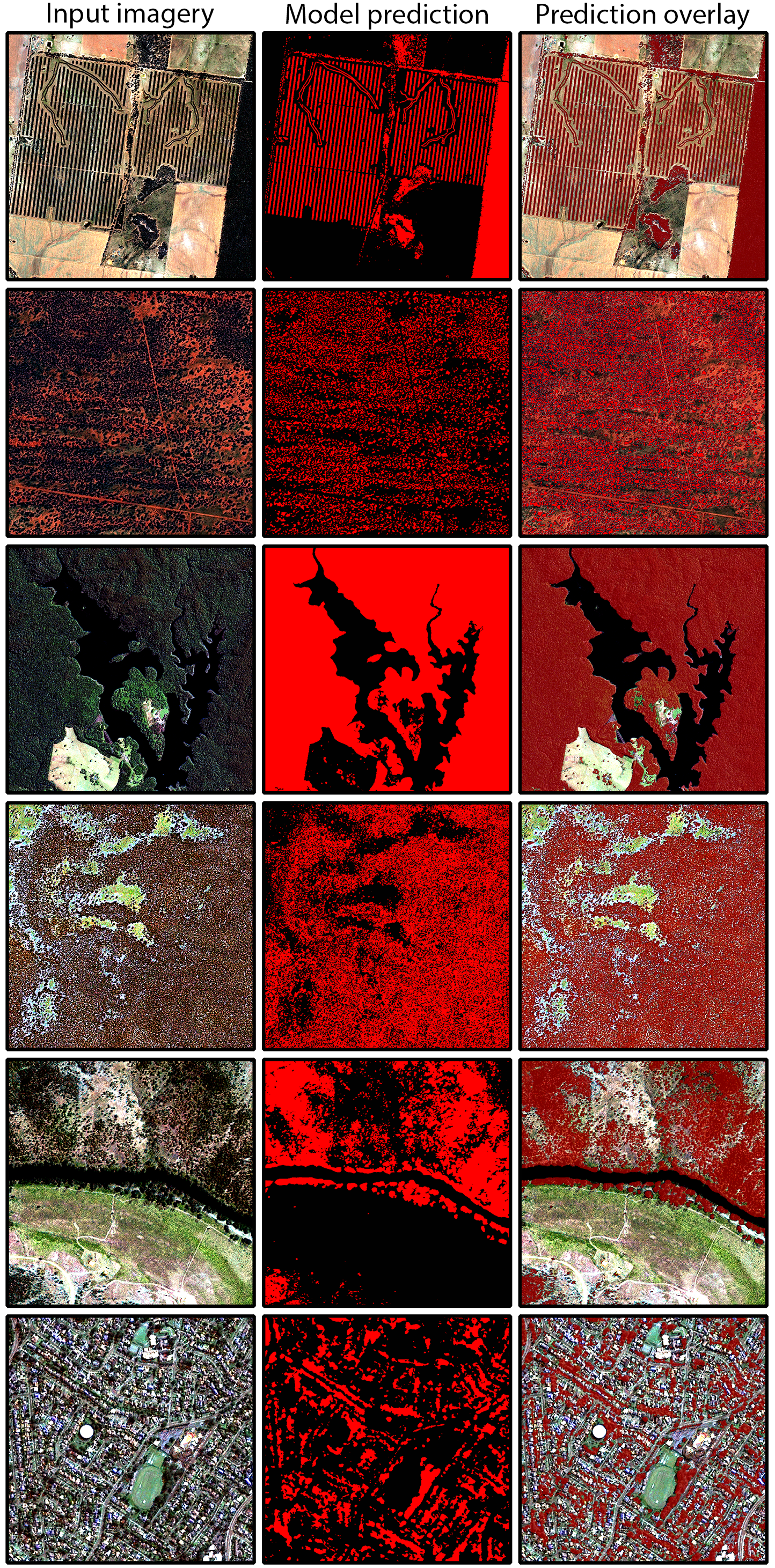}
\caption{Examples of model predictions overlayed with SPOT composite imagery.}
\label{modelPredictions}
\end{figure}

Example model predictions can be seen in Fig.~\ref{modelPredictions}.

\section{Experiment design}
To assess the impact of learning from multiple image sources, an experiment was held where multiple models were trained on various datasets which withheld some of the data sources. A total of 7 models were trained which followed the training procedure outlined in Section~\ref{section_Training}, with the exception of the source images used to construct the dataset. 
4 of the 7 models were trained exclusively on a specific single-date SPOT imagery: 2019, 2020, 2021 \& 2022. The 5th model was trained on all 4 of the single-date SPOT imagery whilst the 6th model was trained exclusively on the SPOT composite imagery. The final model used all available data sources.

\subsection{Validation dataset evaluation}
Each of the 7 trained models were evaluated on each of the 7 validation datasets. 
To rank the performance of each model the weighted overall accuracy (wOA) was used:
\begin{equation}\label{Equation_OA}
\text{wOA} = \frac{\text{wTP} + \text{wTN}}{\text{wTP} + \text{wFP} + \text{wTN} + \text{wFN}},  
\end{equation}
where wTP, wTN, wFP and wFN are the weighted true positives, true negatives, false positives and false negatives respectively which are defined as:
\begin{equation}
\text{wTP} = \sum\limits_{i=1}^{N}{w_i \bar{p_i} y_i},
\end{equation}
\begin{equation}
\text{wTN} = \sum\limits_{i=1}^{N}{w_i (1 - \bar{p_i})(1 - y_i)},
\end{equation}
\begin{equation}
\text{wFP} = \sum\limits_{i=1}^{N}{w_i \bar{p_i} (1 - y_i)}
\end{equation}
and
\begin{equation}
\text{wFN} = \sum\limits_{i=1}^{N}{w_i (1 - \bar{p_i}) y_i}.
\end{equation}
The weighted overall accuracy was used to account for the incompleteness in the labeled image patches and the uncertainty operators have in disambiguating between vegetation above or below 2m in height. 

The results of the validation dataset can be seen in Table~\ref{Table_validation} in Section~\ref{section_results_validation}.

\subsection{Fisher et al. dataset evaluation}
Each of the 7 trained models were evaluated on the labeled point dataset from \cite{fisher2016large} Fisher et al. using the image source in which they were trained on to generate predictions. The \cite{fisher2016large} Fisher et al. dataset contained 6,648 points sampled using a stratified random approach where each point was manually classified through visual interpretation using a Leica ADS40 airborne digital camera at 0.5m resolution captured in 2011.  

Each of the models alongside the \cite{fisher2016large} Fisher et al. model were evaluated on the \cite{fisher2016large} Fisher et al. dataset and were assessed under 3 criteria: woody accuracy, non-woody accuracy and overall accuracy defined as:

\begin{equation}\label{Eq_WoodyAccuracy}
\text{woody accuracy} = \frac{\sum\limits_{i=1}^{N}{\bar{p_i} y_i}}{\sum\limits_{i=1}^{N}{y_i}},
\end{equation}

\begin{equation}\label{Eq_NonWoodyAccuracy}
\text{non-woody accuracy} = \frac{\sum\limits_{i=1}^{N}{(1 - \bar{p_i}) (1 - y_i)}}{\sum\limits_{i=1}^{N}{(1 - y_i)}}
\end{equation}

and

\begin{equation}\label{Eq_OverallAccuracy}
\text{overall accuracy} = \frac{\sum\limits_{i=1}^{N}{\bar{p_i} y_i} + \sum\limits_{i=1}^{N}{(1 - \bar{p_i}) (1 - y_i)}}{N}.
\end{equation}

The results of the \cite{fisher2016large} Fisher et al. dataset can be seen in Table~\ref{Table_fisher} in Section~\ref{section_results_fisher}.

\subsection{Independent point dataset evaluation}
The main limitation of the \cite{fisher2016large} Fisher et al. dataset is the temporal misalignment of image capture dates to that of the SPOT 6/7 source imagery. To attain a more accurate model evaluation that accounts for the temporal change in woody extent, an additional independent point dataset was gathered.
A total of 4,491 points were randomly sampled across 12 of the 50km \(\times\) 50km scenes. 
Each point was manually classified as woody or non-woody through visual interpretation using MAXAR 0.6m imagery.

Each of the models’ predictions were evaluated using the image sources in which they were trained on and were assessed using woody accuracy, non-woody accuracy and overall accuracy which are respectively defined in Equations~\ref{Eq_WoodyAccuracy}, \ref{Eq_NonWoodyAccuracy} \& \ref{Eq_OverallAccuracy}.

To assess the impact that prediction fusion and the SPOT composite has on a model’s ability to generate woody extent maps, the model which was trained on all image sources was tested against different prediction generation methods.
The model was deployed on the individual image sources and SPOT composite, using a confidence threshold of 50\% to convert the confidence values for each image source to a binary woody or non-woody class.

The results of the independent point dataset and image source performance can be seen in Tables~\ref{Table_independent} \& \ref{Table_imageSource} respectively in Section~\ref{section_result_independent}.

\section{Results}\label{section_results}
\subsection{Validation dataset evaluation}\label{section_results_validation}

\begin{table*}
\caption{Weighted overall accuracy on validation datasets}
\hskip 2.5cm
\scriptsize
\label{Table_validation}
\begin{tabular}{l|ccccccc}
\bottomrule
 & \multicolumn{7}{|c}{\textbf{Validation dataset}} \\
\toprule
\textbf{Model} & Single-2019 & Single-2020 & Single-2021 & Single-2022 & All singles & Image composite & All data \\
\midrule
\hline
Single-2019 & 0.979 & 0.806 & 0.796 & 0.827 & 0.854 & 0.872 & 0.861 \\
Single-2020 & 0.889 & 0.977 & 0.943 & 0.934 & 0.935 & 0.923 & 0.931 \\
Single-2021 & 0.819 & 0.958 & 0.983 & 0.950 & 0.925 & 0.956 & 0.936 \\
Single-2022 & 0.799 & 0.917 & 0.929 & 0.972 & 0.901 & 0.902 & 0.902 \\
All singles & 0.978 & \textbf{0.979} & 0.984 & \textbf{0.980} & \textbf{0.980} & 0.970 & 0.976 \\
Image composite & 0.812 & 0.951 & 0.976 & 0.957 & 0.921 & \textbf{0.981} & 0.943 \\
All data & \textbf{0.980} & 0.976 & \textbf{0.986} & 0.976 & \textbf{0.980} & 0.979 & \textbf{0.979} \\
\hline
\bottomrule
\end{tabular}
\newline
Bold values highlight the most optimal value achieved for each dataset.
\end{table*}

The results of the validation dataset evaluation can be seen in Table~\ref{Table_validation}. The best performing models across the different datasets were those which were trained on multiple input imagery sources. The model trained on all data sources and the model trained on all of the single date image sources most often attained the best weighted overall accuracy across the datasets and achieved a weighted overall accuracy of 97.9\% and 97.6\% respectively against all data sources. Compared to the single-date imagery trained models which had an average weighted overall accuracy of 90.7\%, training on multiple image sources was found to reduce the average error by 76.2\%.

When observing the performance of the single-date trained models, it can be seen that the best model performance is attained when evaluating on the dataset with the identical date the model was trained on. 
The 2019 model and dataset was shown to be the most prone to performance degradation, where the model trained on 2019 data saw an average drop of performance from 97.9\% to 81.0\%, a 9\(\times\) increase in the average error across the single-date images.
The drop in performance against the 2019 image source is consistent with the alternative single-date trained models. When comparing the average performance for the 2020-2022 datasets against the 2019 dataset, an average drop in performance from 95.2\% to 83.6\%, a 3.4\(\times\) increase in the average error was observed.

Training on multiple image sources gave the most consistent performance across all of the datasets. For models trained on a single image source, the average standard deviation of model performance across all of the datasets was 0.050.
When compared to the models trained on all of the data and all of the single date images, the average standard deviation across all of the datasets was 0.004, a reduction of 13\(\times\) compared to the single data source models.

\subsection{Fisher et al. dataset evaluation}\label{section_results_fisher}
The model results on the \cite{fisher2016large} Fisher et al. dataset can be seen in Table~\ref{Table_fisher}.
Similarly to the validation dataset evaluation, models which were trained on multiple image sources performed better than those trained on single image sources. The model trained on all data sources and the model trained on all single date image sources performed equally, achieving an overall accuracy of 92.8\%. Compared to single-date imagery trained models which had an average overall accuracy of 89.9\%, training on multiple image sources was found to reduce the average error by 28.1\%.

Despite the 11 years difference in source imagery of which the \cite{fisher2016large} Fisher et al. points were based on, all the trained models outperformed the \cite{fisher2016large} Fisher et al. model.

\begin{table}[b]
\caption{Model performance on Fisher et al. dataset}
\scriptsize
\label{Table_fisher}
\begin{tabular}{l|ccc}
\bottomrule
\textbf{Model} & Woody accuracy & Non-woody accuracy & Overall accuracy \\
\toprule
\hline
Single-2019 & 0.848 & 0.972 & 0.911 \\
Single-2020 & 0.842 & 0.966 & 0.905 \\
Single-2021 & 0.844 & 0.965 & 0.904 \\
Single-2022 & 0.790 & 0.964 & 0.879 \\
All singles & 0.877 & \textbf{0.978} & \textbf{0.928}  \\
Image composite & 0.831 & 0.977 & 0.905 \\
All data & \textbf{0.880} & 0.975 & \textbf{0.928} \\
\hline
Fisher et al. & 0.747 & 0.930 & 0.868 \\
\hline
\bottomrule
\end{tabular}
\newline
Bold values highlight the most optimal value achieved for each metric. 
Model predictions were generated using the source imagery the respective model was trained on.
\end{table}

\subsection{Independent point dataset evaluation}\label{section_result_independent}
The model results on the independent point dataset can be seen in Table~\ref{Table_independent}. The best performing model on the independent point dataset was the model trained on all data sources with an overall accuracy of 97.3\%.
Compared to single-date imagery trained models which had an average overall accuracy of 94.1\%, training on multiple image sources was found to reduce the average error by 46.8\%. 

The results comparing the impact different image sources have on an individual model’s ability to map woody vegetation can be seen in Table~\ref{Table_imageSource}.
The best performing single image source was the SPOT composite with an overall accuracy of 96.4\%, compared to the average overall accuracy across the individual single-date image sources of 94.2\%, resulting in a reduction of 38.2\% in the total error.   
Despite the SPOT composite’s superior performance as a single image source, fusing the predictions across the SPOT composite and individual image sources proved to provide the best results for inference. Fusing predictions from all image sources resulted in an overall accuracy of 97.3\%, a reduction in the average overall error for individual image sources of 53.6\%, or a further 25\% reduction in the error with respect to the SPOT composite.

\begin{table}
\caption{Model performance on the independent point dataset}
\scriptsize
\label{Table_independent}
\begin{tabular}{l|ccc}
\bottomrule
\textbf{Model} & Woody accuracy & Non-woody accuracy & Overall accuracy \\
\toprule
\hline
Single-2019 & 0.806 & 0.978 & 0.936 \\
Single-2020 & 0.806 & 0.981  & 0.938 \\
Single-2021 & 0.850 & 0.980  & 0.939 \\
Single-2022 & 0.874 & 0.980  & 0.952 \\
All singles & 0.889 & 0.990  & 0.965  \\
Image composite & 0.863 & \textbf{0.992}  & 0.960 \\
All data & \textbf{0.914} & \textbf{0.992}  & \textbf{0.973} \\
\hline
\bottomrule
\end{tabular}
\newline
Bold values highlight the most optimal value achieved for each metric.
Model predictions were generated using the source imagery the respective model was trained on.
\end{table}

\begin{table}
\caption{Image source performance on the independent point dataset}
\scriptsize
\hskip -0.5cm
\label{Table_imageSource}
\begin{tabular}{l|ccc}
\bottomrule
\textbf{Image source} & Woody accuracy & Non-woody accuracy & Overall accuracy \\
\toprule
\hline
SPOT 2019 & 0.753 & 0.980 & 0.924 \\
SPOT 2020 & 0.802 & 0.986  & 0.941 \\
SPOT 2021 & 0.882 & 0.979  & 0.949 \\
SPOT 2022 & 0.853 & 0.985  & 0.953 \\
SPOT composite & 0.892 & 0.988  & 0.964  \\
All (prediction fusion) & \textbf{0.914} & \textbf{0.992}  & \textbf{0.973} \\
\hline
\bottomrule
\end{tabular}
\newline
Bold values highlight the most optimal value achieved for each metric.
All predictions were generated using the model trained on all data sources.
\end{table}

\section{Discussion}
\subsection{Reliance on single-date imagery}
The results of the validation dataset evaluation outlined in Section~\ref{section_results_validation} demonstrated a degradation of model performance for models trained on a single date of imagery when being deployed on alternative dates of imagery.
Although such models performed competitively when being assessed on image dates they were trained on, an overall lack of consistency was observed through an average 13\(\times\) increase in the standard deviation in the weighted overall accuracy compared to models trained on multiple image sources. 

Although generating one-off woody extent maps based on a single date may not experience the adverse effects of transferring to alternative image dates, caution needs to be taken when developing a model that is intended to be routinely deployed across time.
When ignoring the 2019 imagery, the average drop in performance of the weighted overall accuracy when transferring to alternative dates was 3.8\%. When compared to the 2019 imagery, the weighted overall accuracy further decreased by an average of 14.2\%.

The large variation in potential model performance degradation makes models trained on image sources derived from a short temporal time span unreliable when extrapolating to unseen time periods. 
By sharing the handcrafted labels derived from the image composite to the individual date imagery, the models trained on multiple image sources were found to be robust to different annual imagery as demonstrated by the 13\(\times\) reduction in the standard deviation of model performance across the datasets.

Compared to the single-date datasets, the multi-date image composition data source was able to recover 7.07\% to 16.53\% of the total pixels impacted by cloud in the training and validation datasets. Resulting in an average of 12.91\% would be missing pixels due to relying on single source imagery resultant from potential cloud cover.   

The degradation of performance from the reliance of single date imagery is further demonstrated in the image source performance evaluation on the independent point dataset outlined in Section~\ref{section_result_independent}.
Inference methods which relied on fusing multiple input sources either by image composition or prediction fusion outperformed inference methods which relied on single-date imagery.
Highlighting the importance of designing models to not be reliant on a single source of imagery, where integrating multiple data sources into all stages of the modelling process positively impacted performance. From the initial training dataset which was shown to improve consistency, to the final output predictions which was shown to increase the accuracy of the generated maps.

However the enhanced performance of incorporating multiple image sources comes at the cost of additional computational resources required. For the prediction fusion method, a 5\(\times\) increase in computational time was required to generate the initial predictions of all 5 image sources compared to a single image source. Although the multi-date image composite approach did not require additional processing during inference, additional pre-processing and artifact storage was required to produce the image composite.

An additional concern of incorporating multiple image sources for label transfer is the potential to introduce label noise from transferring labels defined by the image composite to single-date images. Single-date images may contain cloud or experience changes relative to the image composite due to land clearing which can result in incorrect labels being transferred.

\subsection{Data efficiency}
The main limitation of deep learning approaches is the requirement of large amounts of data to train a model that is sufficiently generalizable and robust during deployment. 
Compared to prior CNN methods shown in Table~\ref{Table_priorwork}, the proposed work was the most efficient in terms of the total labeled pixels required when normalized by the study area.

The data efficiency of the proposed work is credited towards the data augmentation technique of transferring the SPOT composite labels to alternative image sources, multiplying training image diversity.
This data augmentation technique was found to be most influential on model performance as seen in Tables~\ref{Table_validation}, \ref{Table_fisher} \& \ref{Table_independent} in Section~\ref{section_results}.
All models trained on multiple input imagery sources routinely outperformed those trained on a single type of input imagery.  

An additional factor promoting data efficiency was the use of the image patch uncertainty score to propose candidate image patches for labelling as an active learning strategy. By using the fused confidence values generated by an intermediate model, image patches containing difficult or unfamiliar conditions were proposed, allowing for the targeting of specific patches in which the model needed the most improvement. 
An additional benefit of the active learning approach was the ability to attain feedback when sufficient labelled data was collected. It was observed for each consecutive iteration, fewer image patches were proposed, resulting in a greater number of spatially diverse scenes being used to compensate for the lower number of proposed patches. This minimized labelling efforts by ensuring that excessive labels were not collected by stopping the labelling process early when few image patches were proposed from unseen scenes.
However the quantitative impact of the active learning approach is left to future work due to the cost associated procuring additional independent datasets for baseline comparisons against stratified random sampling approaches and the difficulties associated with objectively determining when to stop collecting additional labels.

Despite the proposed work being the most data efficient compared to prior deep learning approaches as shown in Table~\ref{Table_priorwork}, all deep learning CNN methods required orders of magnitude more human annotated examples compared to non-deep learning methods due to containing a larger amount of trainable parameters needing to be optimized. Further work is required to quantify the trade-off of improved performance and generalizability made by deep learning approaches with respect to the data and computational efficiency of non-deep learning methods. Where such future work will be impactful in guiding remote sensing scientists in determining the most practical model choice in terms of cost to performance.

\subsection{Woody vegetation map}
\begin{figure}
\centering
\includegraphics[width=1.0\columnwidth]{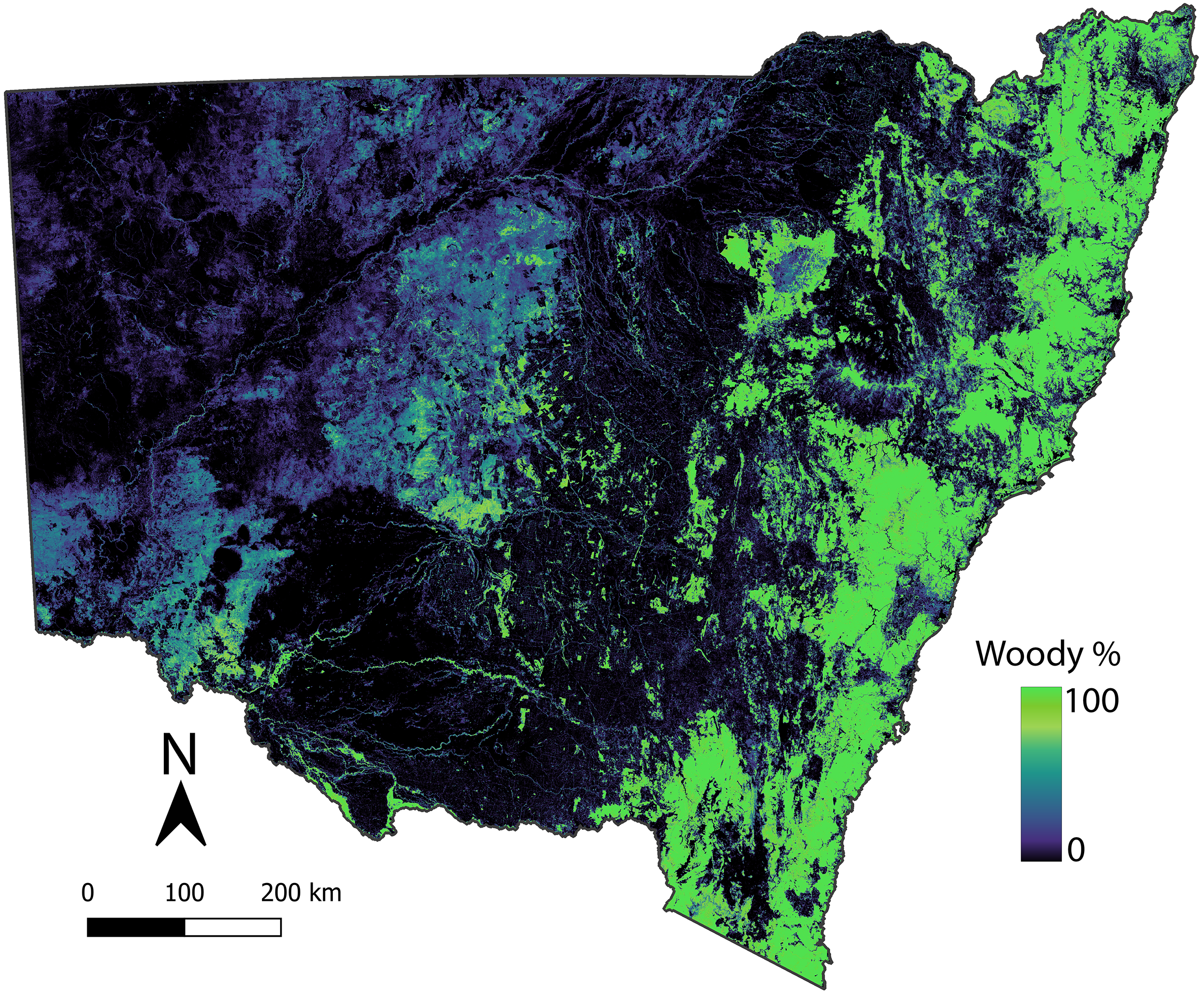}
\caption{Woody vegetation map of New South Wales, Australia. Map was reprojected at 240m pixel resolution using average resampling to attain woody percentage values.}
\label{woodyMap}
\end{figure}

The generated map of woody extent for New South Wales, Australia can be seen in Fig.~\ref{woodyMap}.

\subsection{Historical woody vegetation map comparison}
\begin{figure}
\centering
\includegraphics[width=0.99\columnwidth]{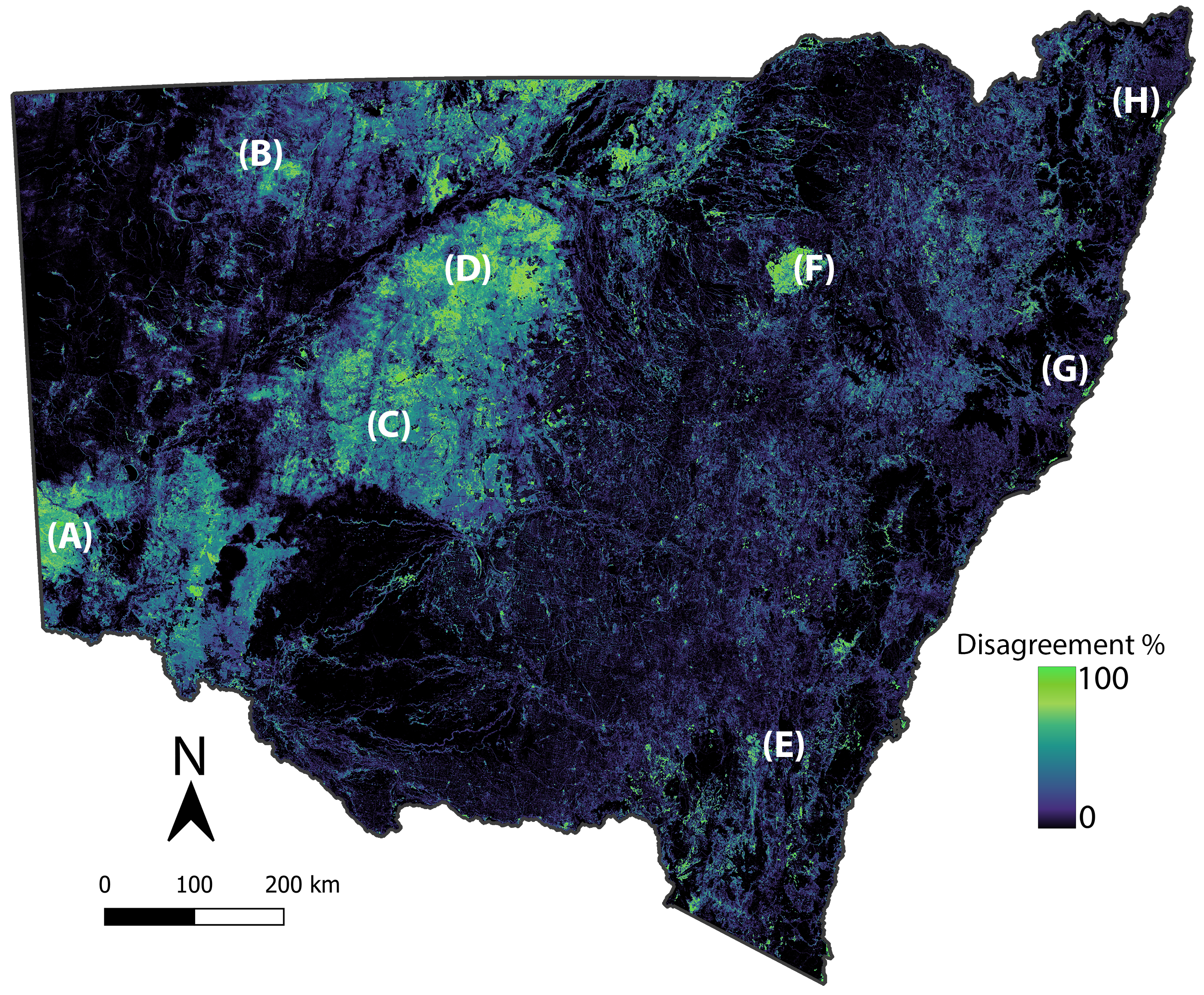}
\caption{Woody classification disagreement map of the proposed framework compared to the \cite{fisher2016large} Fisher et al. map. Markers (A) – (H) indicate regions of interest which are visualized in Fig.~\ref{comparison_images}.}
\label{comparison_map}
\end{figure}

\begin{figure}
\centering
\includegraphics[width=1.0\columnwidth]{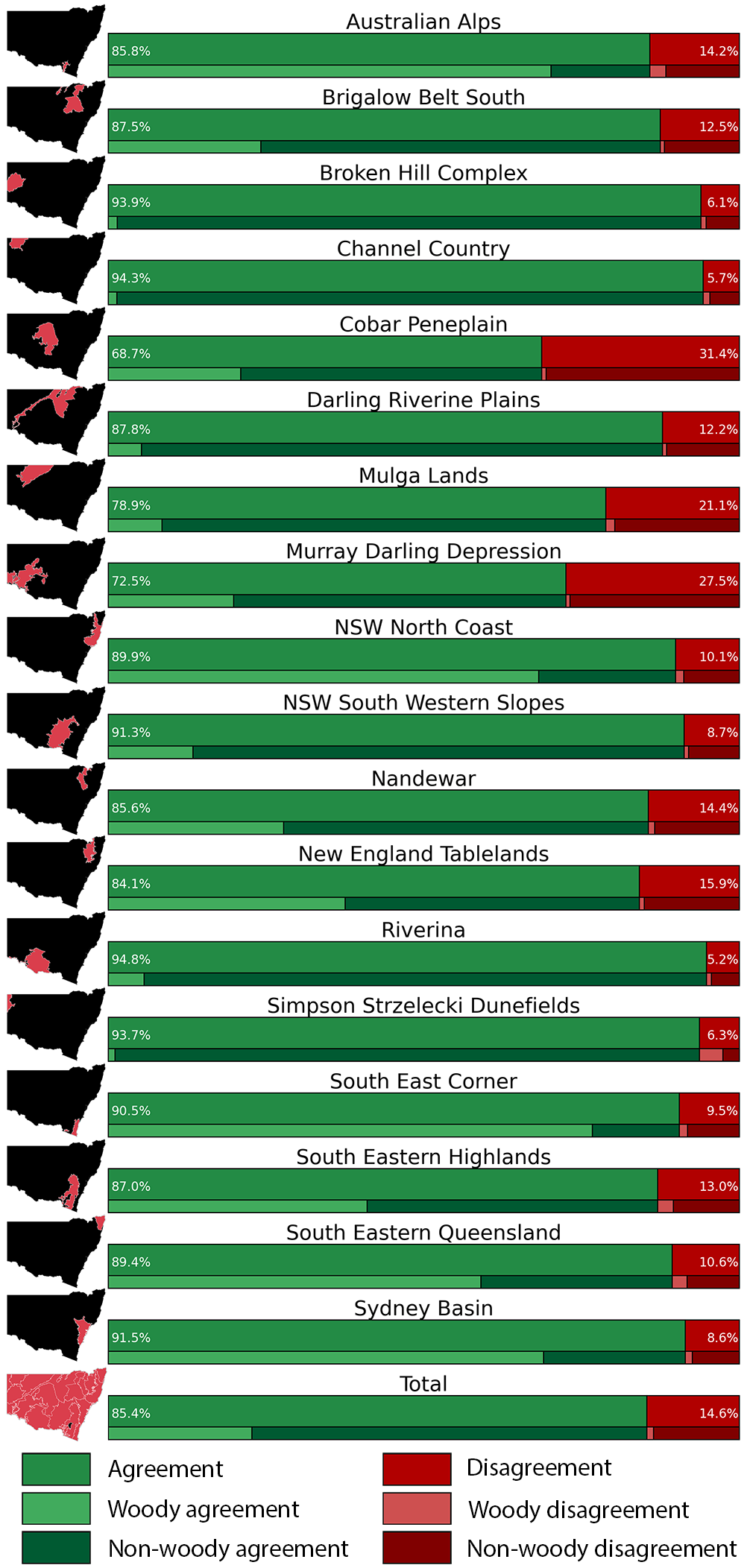}
\caption{Agreement between the proposed framework compared to the \cite{fisher2016large} Fisher et al. map across IBRA bioregions. 
The proposed 1.5m woody vegetation map was reprojected to 5m using mode resampling to match the 5m \cite{fisher2016large} Fisher et al. map.
IBRA bioregions are visualized to the left of the agreement bars.
Agreement is defined as pixels which share the same woody classification between the two maps. Woody disagreement is defined as the proposed model denoting a pixel as woody vegetation whilst the \cite{fisher2016large} Fisher et al. map denoting the pixel as non-woody. Non-woody disagreement is defined as the proposed model denoting a pixel as non-woody whilst the \cite{fisher2016large} Fisher et al. map denoting the pixel as woody vegetation.
}
\label{comparison_ibra}
\end{figure}

The proposed 1.5m woody vegetation map was aggregated using mode resampling to align with historical 5m woody vegetation maps for the study area described in \cite{fisher2016large} Fisher et al., where it was found to share an 85.4\% similarity in classification.
The spatial distribution of woody vegetation classification disagreement can be seen in Fig.~\ref{comparison_map}.
To characterize areas with the highest disagreement, the study area was subdivided into regions using the Interim Biogeographic Regionalisation for Australia (IBRA) \cite{IBRA_2025, IBRA_1995} regions, which describe distinct bioregions based on climate, geology and native vegetation information. The woody vegetation classification disagreement based on IBRA regions can be seen in Fig.~\ref{comparison_ibra}. 

\begin{figure}
\centering
\includegraphics[width=1.0\columnwidth]{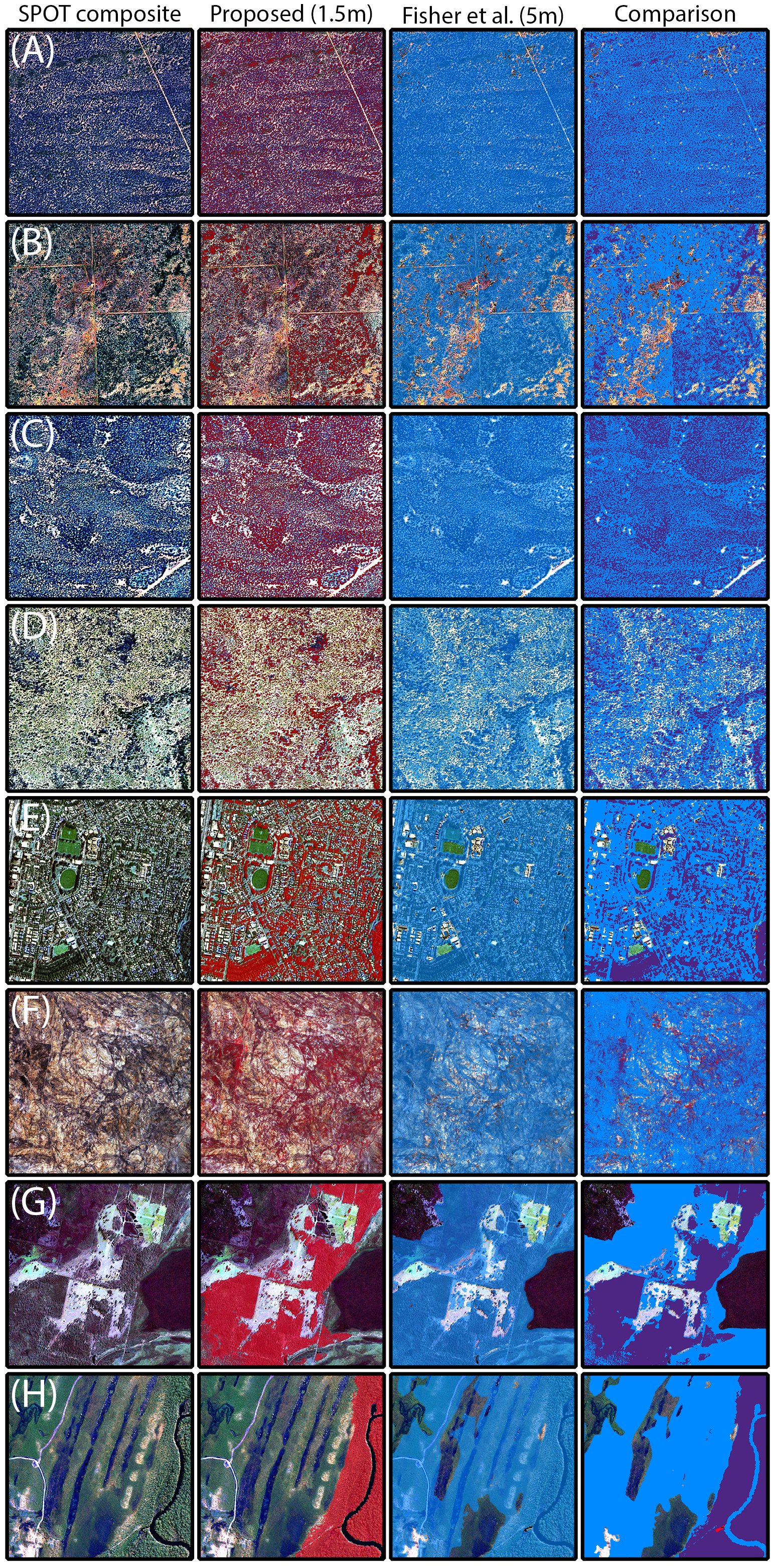}
\caption{Comparison between the predictions of the proposed framework and the \cite{fisher2016large} Fisher et al. model.
Red pixels denote predicted woody vegetation from the proposed method using 1.5m SPOT 6/7 source imagery.
Blue pixels denote predicted woody vegetation from the \cite{fisher2016large} Fisher et al. model using 5m SPOT 5 source imagery.
Purple pixels within the comparison images denote regions of agreement between the two models whilst the red and blue pixels reflect predicted woody vegetation exclusive to their respective model. 
Letters (A) – (H) indicate spatial markers visualized in Fig.~\ref{comparison_map}.}
\label{comparison_images}
\end{figure}

IBRA regions which contained the greatest proportion of woody classification disagreement were the Cobar Peneplain (31.4\%), Murray Darling Depression (27.5\%) and the Mulga Lands (21.1\%). Across the 3 IBRA regions, 96\% of the total disagreement originated from non-woody disagreement, where the proposed worked classified a pixel as non-woody whilst the historical \cite{fisher2016large} Fisher et al. model classified the pixel as woody.
Shared vegetation features across the 3 IBRA regions is the prevalence of woodlands \cite{IBRA_1995}, areas containing widely spaced trees where tree crowns do not touch. 
Observing Fig.~\ref{comparison_images}, IBRA regions Cobar Peneplain (Fig.~\ref{comparison_images}-D \& Fig.~\ref{comparison_images}-C), Murray Darling Depression (Fig.~\ref{comparison_images}-A \& Fig.~\ref{comparison_images}-C) and the Mulga Lands (Fig.~\ref{comparison_images}-B), all share similar non-woody disagreement patterns resultant from woodland vegetation spacing.
The proposed framework individually segmented each tree crown within the sample scenes compared to the historical \cite{fisher2016large} Fisher et al. model which aggregated individual tree crowns into a single homogenous region.
The finer detail of the proposed work’s output can be attributed to the use of higher resolution input imagery of 1.5m compared to 5m, and the greater ability to pool spatial information to detect edge boundaries around tree canopies.
The disparity between individual crown segmentation and region segmentation leads to high non-woody disagreement rates, the most prominent cause of disagreement.

Alternative sources of disagreement result from woody vegetation segmentation around urban areas (Fig.~\ref{comparison_images}-E), temporal misalignment from events such as fire (Fig.~\ref{comparison_images}-F) and the difficulty of classifying woody vegetation around damp coastal regions which contain vegetation close to the 2m woody definition criteria (Fig.~\ref{comparison_images}-G \& Fig.~\ref{comparison_images}-H). 

\section{Conclusion}
In this work we proposed a deep learning framework focused on enhancing data efficiency and robustness against varying image quality to segment woody vegetation defined as vegetation over the height of 2m. SPOT 6/7 imagery was used where a composite image source was proposed by normalizing each SPOT image band by fitting a single and double gaussian before computing the discrete-median for each pixel. 

Labels derived from the SPOT composite were transferred to individual SPOT image sources for enhanced data efficiency, requiring less labels when normalized by study area compared to prior deep learning CNN approaches. Training on multiple image sources was found to lead to the best performing models, capable of reducing errors by up to 76.2\%, whilst showing a greater consistency in performance through a 13\(\times\) reduction in the standard deviation.

For increased accuracy of generated woody extent maps, a multi-image source prediction fusion method was proposed. Generated woody maps derived from composite imagery were found to reduce the overall accuracy error by 38.2\% compared to single-date imagery, whilst a 53.6\% reduction in the overall accuracy error was experienced when performing prediction fusion.

\section*{Acknowledgment}
Computational power was provided by the Science Data and Compute facility from the Science and Insights Division of the New South Wales Department of Climate Change, Energy, the Environment and Water.

\ifCLASSOPTIONcaptionsoff
  \newpage
\fi

\bibliographystyle{IEEEtran}
\bibliography{bibtex/bib/IEEEexample}

\end{document}